# A Foundation Model for DAS Signal Recognition and Visual Prompt Tuning of the Pre-trained Model for Downstream Tasks

Kun Gui, Hongliang Ren, Shang Shi, Jin Lu, Changqiu Yu, Quanjun Cao, Guomin Gu, Qi Xuan

*Abstract*—Distributed Acoustic Sensing (DAS) technology finds growing applications across various domains. However, data distribution disparities due to heterogeneous sensing environments pose challenges for data-driven artificial intelligence (AI) models, limiting cross-domain generalization and facing a shortage of labeled training data. To address these issues, this study proposes a foundational model for DAS signal recognition based on a Masked Autoencoder, named MAEPD. The MAEPD model is pretrained on a dataset of 635,860 samples, encompassing DAS gait spatiotemporal signals, 2D GASF images for perimeter security, 2D time-frequency images for pipeline leakage, and open-dataset signals including whale vocalizations and seismic activities, using a self-supervised mask reconstruction task to capture deep semantic features of DAS signals. Visual Prompt Tuning (VPT) is employed for downstream recognition tasks. This method freezes the pretrained backbone parameters and fine-tunes only a small set of learnable visual prompt vectors inserted into the Transformer encoder layers. Experiments on the NVIDIA GeForce RTX 4080 Super platform validate MAEPD using indoor gait recognition as a downstream task. The VPT-Deep approach achieves a classification accuracy of 96.94% with just 0.322% of parameters fine-tuned, surpassing the traditional Full Fine Tuning (FFT) method by 0.61% and reducing training time by 45%. The model also exhibits robust performance in pipeline leakage detection, confirming the generality, efficiency, and scalability of MAEPD as a foundational model. This approach offers a novel paradigm for addressing the limited generalization of signal recognition models in the DAS domain.

Manuscript received****. This work was supported in part by the National Natural Science Foundation of China (NSFC) (U23A2074, 60907032), in part by the Natural Science Foundation of Zhejiang Province (LZ24F050008, LY20F050009), in part by the Open Fund of the State Key Laboratory of Advanced Optical Communication Systems and Networks, China (2020GZKF013), and in part by the Horizontal projects of public institution, KY-H-20221007, KYY-HX-20210893. The associate editor coordinating the review of this article and approving it for publication was ****. (Corresponding author: Hongliang Ren.)

K. Gui, H. Ren, S. Shi and Q. Xuan are with Institute of Cyberspace Security Zhejiang University of Technology, Hangzhou 310023, China, and Binjiang Institute of Artificial Intelligence, ZJUT, Hangzhou 310056 (e-mail: hlren@zjut.edu.cn);

J. Lu and G. Gu are with the College of Computer Science and Technology, Zhejiang University of Technology, Hangzhou 310023, China;

Q. Cao is with the College of Information Engineering, Zhejiang University of Technology, Hangzhou 310023, China;

C. Yu is with Information Engineering School, Hangzhou Dianzi University, Hangzhou 310018, China.

## I. INTRODUCTION

The phase-sensitive optical time-domain reflectometry (φ-OTDR)-based distributed optical fiber acoustic sensing (DAS) technology detects acoustic or vibrational signals along the fiber by measuring phase changes in the backscattered Rayleigh light caused by light propagation within the fiber [1], [2], [3]. By analyzing the time delay and phase variations of the scattered light, it can accurately locate the source of sound or vibration and extract its characteristics. The DAS technology offers advantages such as distributed sensing, high sensitivity, real-time monitoring, electromagnetic interference immunity, and long-distance coverage. It has been widely applied in fields such as oil and gas resources [4], [5], [6], [7], transportation [8], [9], [10], [11], [12], marine geophysics [13], [14], [15], natural seismic monitoring [16], [17], [18], [19], and perimeter security [20], [21], [22], [23], [24], demonstrating great potential for further applications.

Currently, AI shows great potential in DAS technology [25], [26], [27], covering aspects such as data preprocessing, data augmentation, and event classification. By incorporating AI algorithms, DAS systems can process vast amounts of data more efficiently and intelligently, thereby improving recognition accuracy and decision-making abilities. To enable safe monitoring of oil pipelines using the DAS system, Wu et al. extracted frequency domain features of three events using wavelet decomposition and wavelet packet decomposition methods [28]. They then constructed a four-layer back propagation neural network for event recognition, achieving an identification accuracy of 94.4%. Traditional signal processing and machine learning methods struggle to effectively extract general features from DAS signals. By transforming one-dimensional (1D) DAS vibration signals into two-dimensional (2D) image signals, deep learning models such as convolutional neural networks (CNN) [29], [30], and transformers [31], commonly used in computer vision, can automatically and efficiently identify event features [32], [33]. A study uses an improved convolutional residual block network to extract features from three-dimensional (3D) DAS signals, including time, space, and frequency, achieving high-accuracy event classification [34]. Although the transformative impact of AI models, such as deep learning, on DAS signal classification and recognition is undeniable, many challenges remain to be addressed.

One of the main challenges in DAS applications is the poor generalization ability of AI models across a large number of sensors, primarily due to differences in data distribution among sensors and insufficient training data. Specifically, sensors are typically distributed across various domains with significantly different signal acquisition conditions, such as diverse manufacturing processes and deployment environments. This results in variations in signal distribution, which in turn affects the generalization ability of AI models across different domains [35]. To address this issue, large amounts of labeled data from each domain are typically required for model training. However, when the number of domains is large, this approach becomes cost-inefficient. Therefore, effectively leveraging existing data for model training to enhance generalization across domains has become a key research focus. The available data primarily comes from source and target domains [36]. The target domain refers to the region where the AI model needs to perform tasks, with available labeled data and the ability to automatically acquire unlabeled data. The source domain provides additional labeled data and pre-trained models. Shi et al. propose a full fine tuning (FFT) -based transfer learning approach leveraging the AlexNet network for event classification in novel DAS scenarios [37]. He et al. proposes a weighted partial domain adaptation (PDA) method to address multi-scene sound event classification [38]. This method connects the source and target domains, transferring knowledge from large labeled datasets to small unlabeled datasets, and was successfully applied to fiber-optic perimeter security systems. Sun et al. introduces an adaptive decentralized AI method to improve the generalization performance of DAS signal classification [39]. By fine-tuning a pre-trained model with unlabeled data, this approach trains adaptive AI models for all target domains, significantly reducing false positive rate and false negative rate caused by domain differences. Although researchers have acknowledged the positive impact of pretraining on DAS signal classification, their methods typically involve supervised pretraining for specific tasks. This approach is limited by the scarcity of training data and the challenges of transferring pre-trained weights to other tasks, resulting in poor generalization. Furthermore, significant differences between different DAS signal types hinder the adaptation of pre-trained models to various DAS classification tasks. To address more tasks, researchers must continuously annotate data and improve existing methods, which reduces the efficiency of practical applications [40,], [41]. Therefore, most of these studies lack generalizable model transfer and fail to fundamentally resolve the issue.

A promising solution to address these challenges is to use pre-trained foundational models on a diverse set of DAS signals from various domains [42] [43], [44]. These foundational models are developed through unsupervised pretraining on large amounts of DAS signals, learning the underlying structures and relationships within the data to build a robust knowledge base. Subsequently, the models are fine-tuned with small labeled datasets tailored for specific tasks, ultimately achieving high performance [45], [46]. There are two main training strategies: supervised learning (SL) and self-supervised learning (SSL). SL involves pretraining on labeled data, which is challenging due to the large gap between data annotation speed and data acquisition speed [47]. In contrast, SSL leverages pretext tasks to extract information from large-scale unsupervised data, reducing reliance on labels. Masked autoencoders (MAE) are a typical SSL method [48]. This approach masks part of the input data and uses an encoder-decoder architecture to predict the missing parts, enabling effective representation learning. Through self-supervised pretraining on images, the model learns latent representations, significantly improving the performance of downstream tasks, such as image classification and object detection, even with limited labeled data. In DAS systems, researchers have introduced MAE-based denoising and reconstruction algorithms, significantly enhancing signal quality and improving model's accuracy and efficiency [49], [50]. Recently, He et al. proposed the DAS masked autoencoder (DAS-MAE) framework [51] to learn high-level representations, such as event classes. However, this framework only analyzes 2D spatiotemporal DAS signals and does not establish a general pre-trained model for DAS signals.

This study introduces MAEPD, a pretrained model for DAS signal recognition based on a masked autoencoder. Its performance and versatility were systematically evaluated across DAS signal recognition tasks. The pretraining dataset comprises 635,860 patches of 224 × 224 pixels, including 2D spatiotemporal gait DAS signals, 2D time-frequency DAS data for pipeline leakage vibration events, 1D vibration signals from perimeter security systems converted to 2D images via GAF transformation, and signals from public datasets. MAEPD was pretrained using advanced self-supervised learning with a masked autoencoder (SSL-MAE) and a Vision Transformer-Base (ViT-Base) encoder-decoder architecture, leveraging public and experimentally collected 2D DAS signals from diverse applications. For downstream tasks, Visual Prompt Tuning (VPT) was applied. This approach freezes pretrained model parameters and fine-tunes a small set of visual prompt vectors using limited labeled data from the downstream task. The model was tested in individual identification based on DAS floor-walking patterns, executed on a computer equipped with an NVIDIA GeForce RTX 4080 Super GPU. Results demonstrate that the VPT method achieved a classification accuracy of 96.94% across six gait datasets, outperforming the FFT method by 0.61% and reducing fine-tuning time by 45%. In scenarios like DAS perimeter security and pipeline leakage detection, the VPT-fine-tuned model also exhibited strong classification performance, confirming MAEPD's potential as a versatile foundation for various downstream tasks.

## II. SENSING PRINCIPLE AND SIGNAL PRE-PROCESSING OF DAS

### A. Sensing Principle of DAS

The principal block diagram of the DAS system based on φ-OTDR is shown in Figure 1(a). The system consists of three main components: the sensing fiber, the DAS demodulator, and the signal processing unit. This system measures changes in the vibration state of the fiber and its surroundings by utilizing the backscattered Rayleigh light transmitted through the optical fiber. When external vibrations act on a section of the sensing fiber, the relative position of the Rayleigh scattering centers shifts, causing a local phase change. By detecting the phase variations of the Rayleigh scattering signal, it is possible to accurately capture disturbances such as vibration and sound changes near the fiber.

The working principle is as follows: A narrow-linewidth laser (NLL) emits light, which is split into two paths by a coupler. One path passes through an acousto-optic modulator (AOM), where the light is modulated into pulse form under the control of a modulation signal. After amplification by an erbium-doped fiber amplifier (EDFA), the pulse light enters the sensing fiber through a circulator. The sensing fiber of length $L$ is divided into $N$ segments, with a spatial resolution of $l$, where each segment is considered a discrete reflector. The signals collected from these reflection units are the result of the combined action of $M$ small Rayleigh backscatter centers within the length $L$. As the Rayleigh backscattering propagates along the sensing fiber, the phase delay $\varphi$ can be expressed as $\varphi = \beta L$, where $\beta$ is the propagation constant of light in the fiber. When the fiber at position $z$ is subjected to external disturbances, changes in the fiber length $L$, core diameter $\alpha$, and refractive index $n$ cause variations in the phase of the Rayleigh backscattered light:

$$\Delta\varphi = \beta\Delta L + L\Delta\beta$$
$$= \beta\Delta L + L(\frac{\partial\beta}{\partial n})\Delta n + L(\frac{\partial\beta}{\partial \alpha})\Delta\alpha \quad (1)$$

Here, $\Delta L$, $\Delta n$ and $\Delta\alpha$ represent the phase changes caused by variations in the fiber length, refractive index, and diameter at position $z$, respectively, due to external disturbances.

The backscattered light in the sensing fiber is collected via a circulator and undergoes coherent mixing with the local oscillating light. After I/Q demodulation [52], the phase changes of the coherent Rayleigh backscattered light, which carry vibration information, are received by the photodetector (PD) and converted into digital signals by the analog-to-digital converter (ADC). During each pulse emission cycle, the DAS system acquires an OTDR trace that varies with the spatial position of the fiber. By collecting OTDR traces at different times, a spatiotemporal matrix is constructed to represent the perceived disturbance events in 2D spatiotemporal space. This matrix is used for the detection and analysis of target signals, such as vibrations or sound waves.

*B. Signal Preprocessing of DAS*

OTDR traces may contain background noise, which can interfere with event classification and recognition. Therefore, wavelet denoising preprocessing is applied to the signals obtained by the DAS system. This involves decomposing the 1D time-domain DAS signal into components of different frequencies using wavelet transform, followed by thresholding the wavelet coefficients to remove noise components. Finally, the denoised signal is reconstructed through wavelet inversion. The process is as shown in the following equation:

$$\hat{c}_{ij} = \begin{cases} 0, & |c_{ij}| < \lambda \\ c_{ij}\times(1-\frac{\lambda}{|c_{ij}|}), & |c_{ij}| > \lambda \end{cases} \quad (2)$$

Here, $c_{ij}$ represents the wavelet coefficients, $\lambda$ is the threshold, set to 1 based on empirical values, and $\hat{c}_{ij}$ is the coefficient after thresholding. The coefficients smaller than the threshold are set to zero, while the coefficients greater than the threshold are shrunk before performing inverse wavelet transform to reconstruct the signal. Since vibration events may involve multiple adjacent distributed fiber optic sensing channels, with each channel containing features of the corresponding vibration event, this paper, as shown in Figure 1(b), combines the 1D gait vibration signals from multiple channels into a 2D spatiotemporal image to fully utilize the multi-channel data collected by DAS. In this 2D spatiotemporal image, the horizontal axis represents time (comprising 10,000 sampling points with a total duration of 5 seconds), and the vertical axis represents the channel index (with a total of 7 channels). The subsequent AI-based recognition model is trained and tested based on this 2D spatiotemporal image signal.

## III. PRE-TRAINING OF MAE-BASED FOUNDATION MODEL

A foundation model is a large-scale AI model designed to support multiple tasks. These models are trained on extensive and diverse datasets. This training allows them to adapt to various applications through minimal tuning or task-specific fine-tuning. As general-purpose models, they learn rich linguistic/visual representations from vast amounts of data during a pre-training phase. This paper introduces MAEPD, a foundation model based on the Masked Autoencoders (MAE) framework. MAEPD is pre-trained on a comprehensive collection of 2D Distributed Acoustic Sensing (DAS) data. This dataset includes diverse signal representations such as spatiotemporal, time-frequency, and GAF images from multiple domains. This process enables MAEPD to extract semantic information from various DAS signals accurately. The pre-trained model can then be fine-tuned for different downstream tasks. The primary advantages of the MAEPD foundation model for DAS signal recognition are as follows: (1) Efficiency. Large-scale pre-training reduces the time and computational resources required to develop separate models for each DAS classification task. (2) Generality and Scalability. MAEPD demonstrates strong generality by performing effectively across different tasks and datasets, which also ensures its high scalability. (3) Transfer Learning. The pre-training phase equips MAEPD with powerful feature extraction capabilities. This makes subsequent fine-tuning highly efficient, particularly for tasks with limited labeled data.

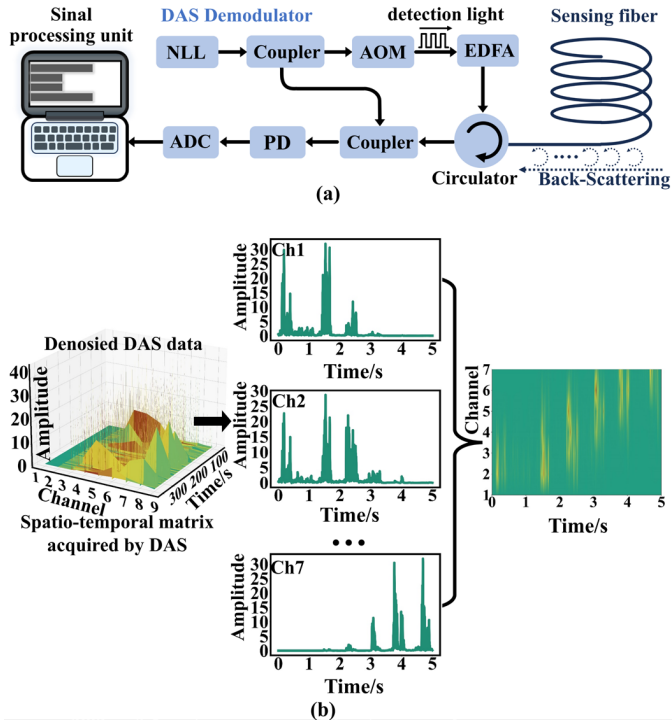

Fig. 1. (a) The main block diagram of the DAS system. (NLL: Narrow Linewidth Laser, AOM: Acousto-Optic Modulator, EDFA: Erbium-doped Optical Fiber Amplifier, PD: Photodetector, ADC: Analog-to-Digital Converter). (b) The signal pre-processing procedure of DAS.

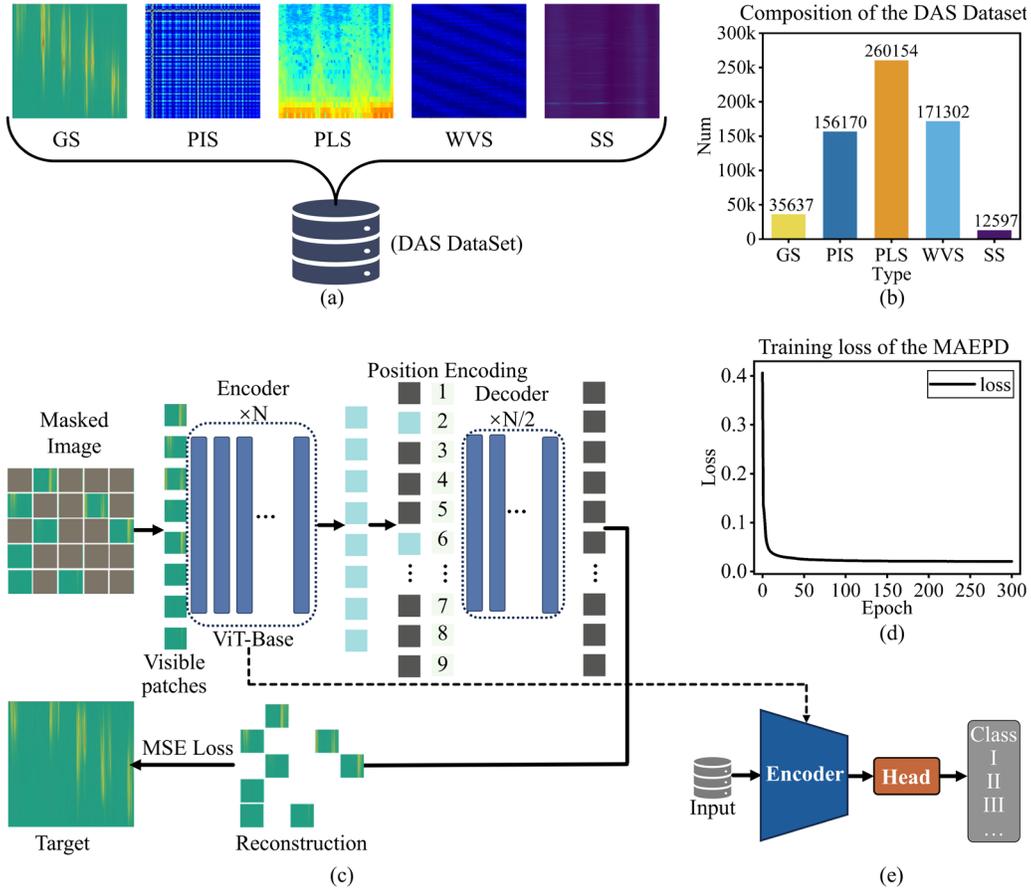

Figure 2. (a) Representative signal samples from the DAS dataset: gait signal (GS), perimeter intrusion signal (PIS), pipeline leakage signal (PLS), and whale vocalization signal (WVS). (b) Composition of the DAS dataset. (d) Training loss curve during the MAEPD pre-training process. (e) Fine-tuning methodology for downstream tasks.

For the MAEPD pre-training phase, a DAS dataset with 635,860 224×224 images was developed to enable self-supervised pre-training of the Foundation model. This dataset includes 2D image data from various DAS applications, as Figure 2(a) depicts. It comprises DAS data from laboratory simulations and publicly available datasets. Spatiotemporal gait data were obtained from walking on carpets or floors in the laboratory, as shown in Figure 1(b). 1D vibration signals from a laboratory DAS perimeter intrusion system were converted into 2D GASF images using the GAF transformation. Similarly, 1D vibration signals from a DAS water pipe leakage setup were transformed into time-frequency images via STFT. The public dataset [53] includes Whale Vocalization and earthquake spatiotemporal images. Sample quantities are presented in Figure 2(b). In the gait recognition scenario, optical fibers placed under carpets captured spatiotemporal DAS signals from experimenters walking in different shoes, yielding 35,637 224×224 gait samples, each with 100,000 sampling points (5 seconds). In the perimeter intrusion scenario, fibers were fixed on fences and the ground to monitor environmental and human intrusion events, producing 156,170 224×224 GASF images. For water pipe leakage, sensing fibers were spirally wrapped around pipe walls to detect micro-vibrations, generating 260,154 224×224 time-frequency DAS images, each with 2,000 sampling points (1 second). Whale signal samples from the University of Washington's Karrenbach team [54], collected at the OOI off Pacific City, Oregon, included 171,302 598×434 spatiotemporal images, each with 60,000 sampling points (60 seconds). Earthquake data from Pennsylvania State University, with fibers buried 1–10 meters underground in the Appalachian Valley [55], comprised 12,597 598×434 spatiotemporal images, each with 300,000 sampling points (600 seconds). Additionally, a perimeter security DAS dataset from Beijing Jiaotong University's Yu Kuanglu team provided 15,418 224×224 images [56].

Figure 2(c) depicts the MAEPD framework, which utilizes an asymmetric encoder-decoder architecture for efficient self-supervised learning and feature learning via masked image reconstruction. The framework comprises an encoder and a decoder. The encoder, based on a ViT, processes only visible, unmasked patches to extract deep semantic features. The decoder reconstructs the complete image from latent representations. Both components are built from stacked transformer encoder blocks, each consisting of alternating multi-head self-attention (MHA) layers and MLP blocks. Layer normalization is applied before each MHA layer and MLP block. Residual connections are incorporated after each MHA layer and MLP block.

The MAEPD framework is outlined as follows: each input DAS signal image is divided into $m$ fixed-size patches: $\{I_j \in \mathbb{R}^{3 \times h \times w} | j \in \mathbb{N}, 1 \leq j \leq m\}$, where $I_j$ represents each patch of the image sample. Here, $h$ and $w$ refer to the height and width of each patch, respectively. Visible patches are represented by $I_j^v$, while randomly masked patches are represented by $I_j^n$. The visible patches $I_j^v$ are then embedded into a $d$-dimensional space:

$$e_0^j = \text{Embed}(I_j^v) \quad e_0^j \in \mathbb{R}^d, j = 1, 2, ...m \quad (3)$$

$$E_i^v = \{e_i^j \in \mathbb{R}^d | j \in \mathbb{N}, 1 \leq j \leq m\} \quad (4)$$

Here, $e_0^j$ represents the $d$-dimensional embedding of the $j$-th visible image patch. $E_i^v$ denotes the collection of $d$-dimensional embeddings for all visible patches in the $i$-th encoder layer. A class token is concatenated with the patch embeddings, followed by the addition of positional encoding to retain positional information. This forms the complete input sequence for the transformer Encoder, which is then fed into $N$ layers of the transformer Encoder:

$$[x_i, E_i^v] = L_{Ei}([x_{i-1}, E_{i-1}^v]) \quad i = 1, 2, ..., N \quad (5)$$

Here, $x_i \in \mathbb{R}^d$ represents the $d$-dimensional embedding of the class token in the $i$-th encoder layer. $[x_i, E_i^v]$ is the concatenation of $x_i$ and $E_i^v$ along the first dimension, i.e., $[x_i, E_i^v] \in \mathbb{R}^{(1+m) \times d}$, After processing through N layers of the transformer encoder, the encoder produces a semantically rich representation vector $E_N^v$ for the visible patches. This vector is concatenated with masked tokens: $E^n$, in the original patch order to form the sequence $[E_N^v, E^n]$. To retain the spatial position of each token in the original image, positional encoding is reapplied to the sequence, which includes encoded visible patches and masked tokens. The decoder processes this sequence, and its output is mapped to the original pixel space via a linear layer to reconstruct the image. The loss function, computed on the masked patches, is optimized by minimizing the mean squared error (MSE) between reconstructed and original pixels:

$$MSE = \frac{1}{\mathcal{M}} \sum_{i \in \mathcal{M}} \|x_i - \hat{x}_i\|^2 \quad (6)$$

Here, $\mathcal{M}$ represents the set of pixels in the masked region. $x_i$ and $\hat{x}_i$ denote the pixel values at pixel $i$ in the original and reconstructed images, respectively. The loss is computed solely in the masked region, allowing the model to concentrate on this challenging predictive task.

During MAEPD pre-training, all input images are resized to 224×224 pixels via random cropping, with a 75% random mask ratio [48]. The AdamW optimizer is employed with a learning rate of 0.001 and a cosine annealing schedule. Training uses a batch size of 16 for 300 epochs on an Intel Core i5-12600KF and NVIDIA GeForce RTX 4070 Ti, requiring 360 hours and 20 minutes. Figure 2(d) illustrates the pre-training loss curve. Rapid convergence occurs in the initial phase (first 25 epochs), with the loss decreasing sharply from 0.40. This reflects efficient learning of the data's underlying structure and key features for reconstructing masked regions. In the middle phase (epochs 25–50), the learning rate slows, and the loss declines steadily. After 200 epochs, the loss curve stabilizes near 0.020, indicating full convergence. This confirms the MAE model's superior DAS image reconstruction capability and accurate extraction of semantic information. Post pre-training, the lightweight decoder is discarded, and the trained encoder is retained for feature extraction or fine-tuning in downstream visual tasks, as depicted in Figure 2(e).

## IV. VISUAL PROMPT TUNING OF THE PRE-TRAINED MODEL FOR DOWNSTREAM TASKS

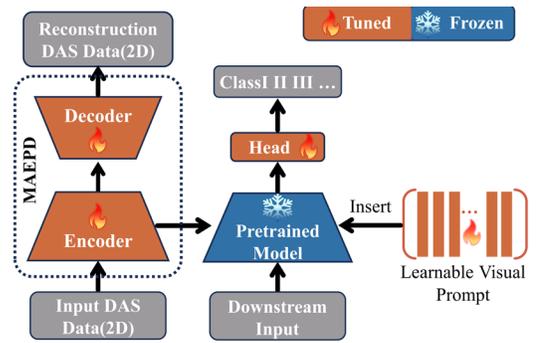

Fig. 3. Visual Prompt Tuning Based on MAEPD.

To adapt the Foundation model for downstream classification tasks, a classification head is added to the pre-trained encoder, followed by fine-tuning for specific tasks. Traditional FFT has proven effective for downstream tasks. However, as network models scale up and the number of parameters grows, FFT requires adjusting most or all model parameters. This leads to parameter updates on the order of millions or even billions. This process involves computing the gradients for the entire model, resulting in high GPU memory usage and significant consumption of computational resources and time. Additionally, extensive parameter updates may undermine the generalization ability of the pre-trained model, leading to the degradation of knowledge from upstream tasks. To address these challenges, this study employs VPT [57]. Similar to prompt tuning in natural language processing, VPT guides pre-trained visual models to learn task-specific objectives for the downstream tasks through the design of input prompt vectors [58], [59], rather than directly modifying all model weight parameters. Due to the strong heterogeneity and dynamic time-varying characteristics of the DAS system environment, data-driven models trained in specific scenarios generally lack cross-domain generalization and adaptability. The VPT process is illustrated in Figure 3. Initially, the MAEPD encoder is frozen as the backbone network for downstream tasks. Next, several learnable visual prompt vectors are designed and inserted into the input feature sequence. The visual prompt vectors are fine-tuned based on a small labeled DAS image dataset. Finally, under the premise of freezing the core backbone network, the number of visual prompt vectors is optimized to ensure high adaptability of the model to downstream tasks. Testing is conducted using the fine-tuned visual prompt vectors with the optimized number and the frozen backbone network.

### A. Design and Insertion of Visual Prompt Vectors

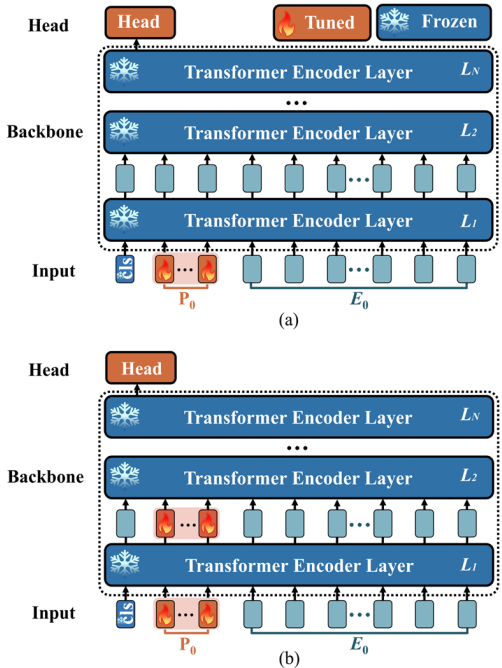

Fig. 4. Visual prompt vectors are incorporated into the Encoder of MAEPD architecture to adapt it to a downstream task. (a) VPT-Shallow: Several visual prompt vectors are added only to the input of the first transformer encoder layer. (b) VPT-Deep: Several visual prompt vectors are added to the input of each transformer encoder layer.

In the MAEPD framework, the pre-trained encoder introduces a set of continuous embedding vectors, or prompt vectors, into the input space. These vectors have the same dimensionality as the image patch embeddings ($d$-dimensional). During fine-tuning, only the task-specific prompt vectors are updated, while the encoder backbone remains frozen. Based on the number of transformer encoder blocks involved, two variants exist: shallow visual prompt tuning (VPT-Shallow) and deep visual prompt tuning (VPT-Deep), as illustrated in Figure 4.

Figure 4(a) shows the VPT-Shallow structure based on the MAEPD encoder. For this VPT variant, several visual prompt vectors are added only to the input of the first transformer encoder block. Each prompt vector is a learnable $d$-dimensional vector. The set of $p$ prompt vectors is represented as:

$$P_0 = \{p^k \in \mathbb{R}^d \mid k \in \mathbb{N}, 1 \leq k \leq p\} \tag{7}$$

The input-output relationship of each transformer encoder block in the VPT-Shallow is as follows:

$$[x_1, \_\_, E_1] = L_1([x_0, P_0, E_0]) \tag{8}$$

$$[x_i, E_i] = L_i([x_{i-1}, E_{i-1}]) \quad i=2,3,...N \tag{9}$$

$$y = \text{Head}(x_N) \tag{10}$$

Here, the set $P_0 \in \mathbb{R}^{p \times d}$ represents the $p$ prompt vectors added to the input. Furthermore, $[x_i, E_i] \in \mathbb{R}^{(1+m) \times d}$, A multilayer perceptron head (Head) is then used to map the class token $x_N$ from the $N$-th layer to the predicted class probability distribution y.

Figure 4(b) illustrates the VPT-Deep architecture based on the MAEPD encoder. In this approach, learnable prompt vectors are inserted into the input of each transformer encoder block. Specifically, for the ($i$+1)-th transformer encoder block $L_{i+1}$, a set of prompt vectors is added to its input:

$$P_i = \{p_i^k \in \mathbb{R}^d \mid k \in \mathbb{N}, 1 \leq k \leq m\} \quad i=0,1,2,...,N \tag{11}$$

The input-output relationship of each transformer encoder block in the VPT-deep is as follows:

$$[x_{i+1}, \_\_, E_{i+1}] = L_{i+1}([x_i, P_i, E_i]) \quad i=0,1,2,...,N \tag{12}$$

Adding prompt vectors to each transformer encoder layer increases the input dimension, which in turn expands the output dimension. Since prompt vectors are introduced in every layer from the first to the last, the output dimension would grow with the number of layers. Therefore, as shown in Equation (11), to maintain a consistent input dimension across layers, the prompt vectors are removed from the output after each Transformer Encoder layer. In addition to the VPT-Deep and VPT-Shallow variants, prompt vectors can also be inserted into the inputs of intermediate transformer encoder layers in ViT (e.g., layers 1 to 3 or layers 9 to 12) [58].

### B. Fine-tuning Based on Visual Prompt Vectors and Optimization of Their Quantity

After the design and insertion of visual prompt vectors, this section presents fine-tuning based on visual prompts and optimization of their quantity to ensure high adaptability to downstream tasks. For downstream classification task, fine-tuning is performed using the cross-entropy loss function:

$$\text{Loss} = -\frac{1}{N} \sum_{n=1}^{N} \sum_{c=1}^{M} y_c^{(n)} \log(\hat{y}_c^{(n)}) \tag{13}$$

Here, $N$ denotes the number of samples, $M$ the number of classes, $y_c^{(n)}$ the true probability distribution of $n$-th sample belonging to class $c$ and $\hat{y}_c^{(n)}$ the predicted probability distribution of $n$-th sample for class $c$.

Based on a small amount of labeled data from the downstream tasks, fine-tuning is performed by computing the loss function and updating the visual prompt vectors using stochastic gradient descent (SGD), aiming to improve prediction performance on downstream task samples. Since all parameters of the pretrained model are frozen, only the learnable prompt vectors and the classification head are updated and stored. In contrast, FFT modifies and stores nearly all model parameters.

For example, given a ViT-Base encoder model with 85,803,270 parameters, where the patch embedding dimension is $d$=768, and 30 shallow or deep prompt vectors are used, the additional parameters are: VPT-shallow: $p×d$=30×768=23,040 and VPT-deep: $N×p×d$= 12×30×768=276,480. Here, $p$ is the number of prompt vectors, $d$ is the embedding dimension, and $N$ is the number of transformer encoder blocks in the ViT-Base encoder. Fine-tuning with VPT-Shallow and VPT-Deep involves only 0.027% and 0.322% of the total ViT-Base parameters, respectively, leading to significant reductions in memory usage and computational cost compared to the FFT method. During inference, the preprocessed DAS 2D signal is concatenated with the fine-tuned prompt vectors and fed into the frozen pretrained model. The final classification is performed using the MLP head. In this work, the number of prompt vectors is also optimized to achieve the best transfer learning performance.

## V. Experiment and Downstream Task Dataset

### A. Experimental Scenes

Gait is a unique biometric feature that reflects the movement patterns of the body during walking, including stride length, cadence, step width, and gait cycle. Gait recognition analyzes these dynamic patterns to identify individuals or monitor behavior, with broad applications in biometrics, health monitoring, and security [60]. Zhou et al. [61] deploy optical cables along a corridor to detect pedestrian gait signals using a DAS system and apply a convolutional long short-term memory network (ConvLSTM) for gait recognition. The experimental results show that the proposed ConvLSTM network achieves an accuracy of 81.35% in distinguishing footstep events from background noise and other non-footstep events. This method demonstrates the superiority and potential of applying DAS technology for gait recognition in large-area scenarios. Shi et al. [62] deploy optical fibers in the ground layer to collect pedestrian gait signals using a DAS system and apply a dual YOLO model to analyze stride length and cadence features for footstep event identification and localization. This method, for the first time in the DAS field, achieves identity recognition of three pedestrians with an accuracy of approximately 86.0%. This study employs gait vibration signals captured by DAS devices for individual identity recognition to evaluate the effectiveness of the MAEPD foundation model and VPT.

As shown in Figure 5(a), an indoor gait signal collection experimental setup is constructed using DAS technology. Fiber optics, encased in protective sleeves, were laid on the back surfaces of a floor, with a person walking over them. The DAS system is used to collect gait data. In this experiment, the dataset obtained from walking on the floor is set as the downstream task dataset. The fiber optics is connected to a DAS demodulator, with a detection range of 5 km, 960 channels, a spatial resolution of 5 m, a gauge length of 5 m, and a signal sampling rate of 20 kS/s. The sensing fiber used is a G652.D single-mode fiber with a protective sleeve, and the fiber connectors are FC/APC. An optical fiber is coiled from the starting point of the DAS demodulator to a length of 2 km, and then connected to the back surfaces of the floor. Figure 5(b) illustrates the actual setup of the floor's front and back surfaces and the fiber layout. The fiber was laid in an "S" shape along its shorter edge on the back of the floor, positioned above a 3 cm high floor joist, and fixed with glue. The fiber layout on the floor covers an area of 1 m × 2 m, with 5 cm spacing between adjacent two parallel optical fibers. The total length of fiber laid on the floor's back is approximately 40 m (8 channels).

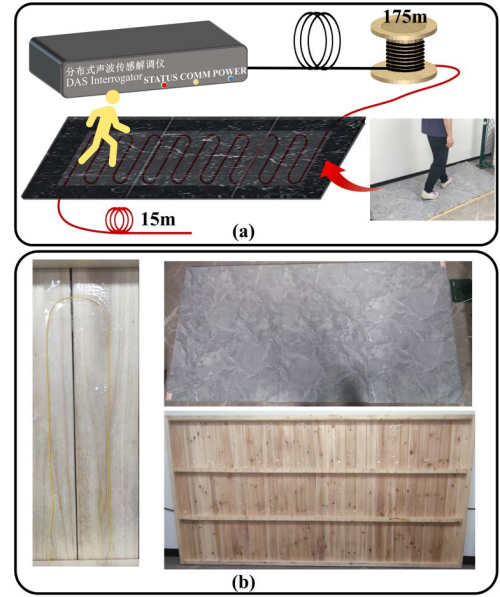

Fig. 5. (a) Experimental setup for collecting gait signals using DAS technology. An optical fiber is coiled from the DAS demodulator's starting point, extending over 2 km, before being connected to the back surfaces of both the floor. (b) Actual images showing the front and back of the floor with the optical fiber arrangement.

### B. Overview of the downstream task dataset

The DAS gait data collected from walking on the floor is used as the downstream task dataset and was not utilized in upstream model training. Multi-channel gait data was collected using the DAS system from three testers wearing different types of shoes (sneakers and slippers) while walking on the floor. Tester I is 178 cm tall, weighs 75 kg, and wears size 42 shoes; Tester II is 175 cm tall, weighs 80 kg, and wears size 41 shoes; Tester III is 170 cm tall, weighs 60 kg, and wears size 40 shoes. A total of 6 types of data samples have been collected, corresponding to the walking data of testers I, II, and III in slippers and sneakers on the floor. As previously mentioned, the collected gait data is denoised using wavelet transformation, and 1D vibration data from the corresponding eight channels within a 5-second duration is combined into 2D DAS spatiotemporal image samples. The dataset details are shown in Table I: 2,520 image samples (420 samples per class across six classes), with 80 samples per class for the training set, 40 for the validation set, and 300 for the testing set.

TABLE I
CLASSIFICATION OF DAS GAIT SIGNAL EVENTS AND CORRESPONDING DATASET DETAILS

| Type | Event | Class | Train/Val/Test | Num |
|---|---|---|---|---|
| Downstream Task Dataset | Tester I-Slippers | I | 80/40/300 | 420 |
| | Tester I-Sneakers | II | 80/40/300 | 420 |
| | Tester II-Slippers | III | 80/40/300 | 420 |
| | Tester II-Sneakers | IV | 80/40/300 | 420 |
| | Tester III-Slippers | V | 80/40/300 | 420 |
| | Tester III-Sneakers | VI | 80/40/300 | 420 |

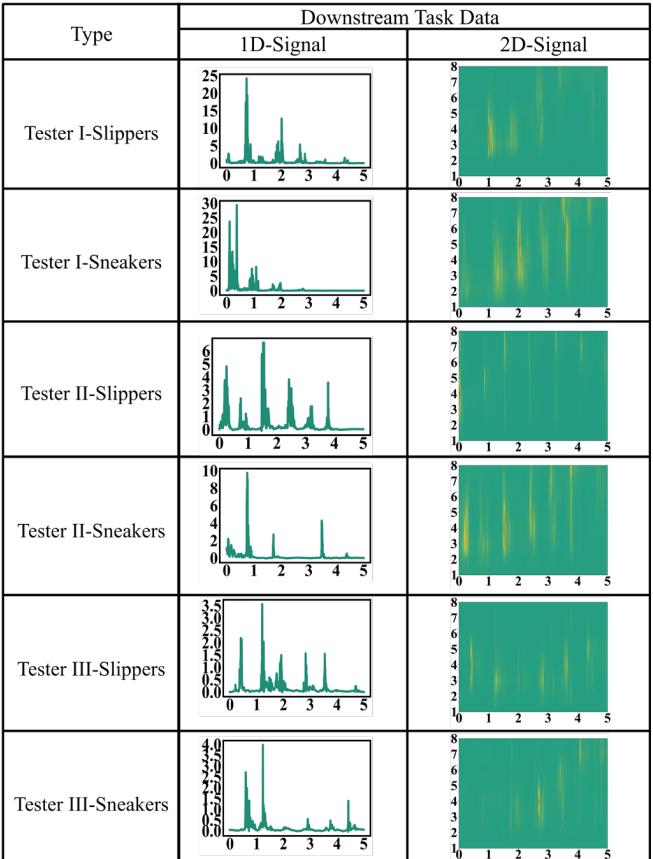

Fig. 6. The 1D gait vibration signals and multi-channel 2D spatiotemporal signals of three testers walking in slippers and sneakers on floor.

Figure 6 shows the 1D gait vibration signals and multi-channel 2D spatiotemporal signals of three testers walking in slippers and sneakers on the floor. When walking on the floor, the pressure is transmitted indirectly through the floor to the sensing fiber, protecting it from damage. The analysis reveals that factors such as body weight and shoe size directly influence the strength of the gait signals and their distribution across channels. Heavier participants generate stronger signals. For the same tester, wearing slippers, which are softer than sneakers, results in the vibration signals being partially buffered, leading to slightly weaker signals compared to when wearing sneakers.

VI. RESULT AND DISCUSSION

A. Fine-tuning for downstream task

For floor gait recognition in downstream tasks, the pre-trained MAEPD encoder is fine-tuned using a small labeled dataset. All parameters of the MAEPD encoder are frozen, and learnable visual prompt vectors are introduced into the input for fine-tuning. For comparison, traditional transfer learning algorithms, FFT and linear probing (LP), are also applied to fine-tune the model. The FFT method adjusts all the weight parameters of the pre-trained model using the limited labeled data from the downstream task, while the LP method keeps the feature extraction layers of the pre-trained model fixed and only trains a linear classifier (usually a fully connected layer) [63]. Table II presents the hyperparameter settings and optimization ranges for different fine-tuning methods. A fixed random seed is used to ensure reproducibility. To find the optimal hyperparameters, learning rates, and weight decay values for the specific task, grid search is performed using the validation set. Following the linear scaling rule, the learning rate is set as base_lr×batch_size/256, with batch_size set to 16, and base_learning rate (lr) selected from the predefined range in Table II. For the FFT and LP methods, the base learning rate range is set to [0.0001, 0.005], while the base_lr range for the proposed VPT method is set to [0.05, 5]. During training, an image augmentation strategy is employed: gait images in the downstream task are first cropped to a large size of 256×256, then randomly cropped to 224×224, followed by random horizontal flipping. Finally, the images are normalized using the mean and standard deviation of ImageNet. All models are implemented using the PyTorch deep learning framework and executed on a personal PC with an Intel Core i9-14900KF CPU and an NVIDIA GeForce RTX 4080 Super GPU.

TABLE II
HYPERPARAMETER OPTIMIZATION OF DIFFERENT FINE-TUNING METHODS FOR GAIT RECOGNITION ON THE TARGET DOMAIN

| Method | FFT or LP | VPT |
|---|---|---|
| Optimizer | SGD | SGD |
| Optimizer momentum | 0.9 | 0.9 |
| base_lr range | [0.0001,0.005] | [0.05,5] |
| Weight decay range | [0,0.01] | [0,0.01] |
| Learning rate schedule | Cosine decay | Cosine decay |
| Total epoch | 200 | 200 |

B. Performance comparison of VPT based on different visual prompt vector insertion methods

This study evaluates the performance of VPT-Deep and VPT-Shallow for gait recognition in downstream tasks. As shown in Table I, the dataset for fine-tuning consists of 480 2D spatiotemporal DAS gait signal samples (80 samples per class from six classes), the validation set consists of 240 signal samples (40 samples per class from six classes), and the test set contains 1800 samples (300 samples per class from six classes). Figure 7(a) shows the performance of VPT-Deep and VPT-Shallow as a function of the number of inserted visual prompt vectors. The results indicate that the accuracy of both methods increases with the number of prompts. However, VPT-Shallow reaches its highest accuracy of 87.56% when the prompt number increases to 25, and slightly decreases to 86.17% when the number of prompts reaches 30. This may be due to an excessive number of visual prompt vectors introducing too many learnable parameters or redundant information, leading to overfitting and reducing generalization ability on the downstream task test data. In contrast, VPT-Deep demonstrates significantly higher accuracy at most prompt lengths. With a prompt number of 1, VPT-Deep achieves a classification accuracy of 89.67%, which rises to 96.94% when the prompt count reaches 30, showing higher stability and superior performance. This suggests that VPT-Deep, by distributing prompt vectors across multiple encoder layers, can more effectively leverage the increased prompt information, enhancing the model's ability to capture complex task features and achieving higher accuracy and robustness in downstream tasks. Therefore, VPT-Deep exhibits stronger expressive and generalization abilities in MAEPD-based downstream tasks.

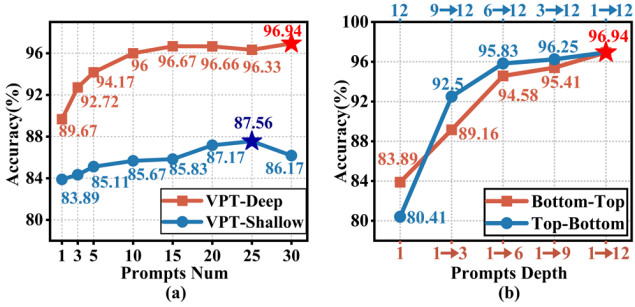

Fig. 7. (a) The performance of VPT-Deep and VPT-Shallow as a function of the number of inserted visual prompt vectors. (b) The performance of both strategies (bottom-top and top-bottom) as a function of the insertion depth of these prompts. In the bottom-top strategy, prompts are added from the first encoder layer onward; in the top-bottom strategy, prompts are added starting from the last encoder layer and moving backward.

We have also investigated the impact of adding prompt vectors to different MAEPD encoder layers on downstream task performance. Specifically, we consider two prompt insertion strategies: bottom-top and top-bottom. In the bottom-top strategy, prompts are added from the first encoder layer onward; in the top-bottom strategy, prompts are added starting from the last encoder layer and moving backward. When the number of visual prompt vectors input to each encoder layer is 30, Figure 7(b) shows the performance of both strategies as a function of the insertion depth of these prompts. The results indicate that, under both strategies, as the number of encoder layers receiving prompts increases, the model's accuracy is improved. In the bottom-top strategy, when prompts are added only to the first layer, the accuracy is 83.89%. As more layers receive prompts, accuracy significantly increases, reaching the maximum value of 96.94% when prompts are added to all layers (1-12 layers), corresponding to VPT-Deep. Similarly, the top-bottom strategy achieves 96.94% accuracy with prompts in all layers (1–12), also corresponding to VPT-Deep. The top-bottom strategy generally outperforms the bottom-top approach. In conclusion, this study adopts the prompt insertion strategy of VPT-Deep and optimizes the number of prompts, achieving the best performance in downstream gait recognition tasks.

### C. Impact of the number of added visual prompt vectors on downstream task performance

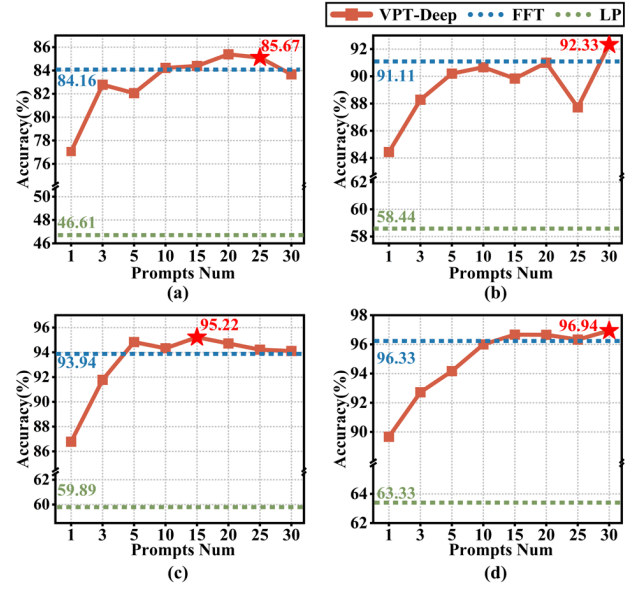

Fig. 8. the test accuracy changes as a function of the number of prompt vectors added, for VPT-Deep, FFT, and LP methods, with downstream task training data sizes of (a) 120, (b) 240, (c) 360, and (d) 480.

This study examines the impact of the number of visual prompt vectors added to each layer in the VPT-Deep variant on downstream task performance. Insufficient prompt vectors may limit model expressiveness, while excessive vectors can introduce redundancy, leading to overfitting and reduced generalization. The optimal prompt vector count varies with training data size. Figures 8(a)-(d) separately show the test accuracy changes as a function of the number of prompt vectors added, for VPT-Deep, FFT, and LP methods, with downstream task training data sizes of 120, 240, 360, and 480. Since FFT and LP are independent of the number of prompt vectors, their test accuracy remains unchanged regardless of the number of prompts. In contrast, VPT-Deep improves performance with a larger number of prompts. When the target domain training data is 120, VPT-Deep reaches the highest accuracy of 85.67% with 25 prompt vectors. With 240 samples, the highest accuracy is 92.33% with 30 prompt vectors. When the downstream task training data increases to 360, adding 15 prompt vectors leads to a continuous increase in accuracy, reaching 95.22%, and adding 20 prompts slightly reduces the accuracy to 94.72%. With 480 downstream task samples, VPT-Deep achieves the highest classification accuracy of 96.94% with 30 prompt vectors. Overall, with smaller training data sizes (120 or 240), the accuracy increases initially and then decreases as more prompts are added, showing considerable fluctuation. However, with larger training data sizes (480 or 360), the accuracy increases steadily with more prompts and shows less fluctuation. Notably, when the training data consists of 480 samples and 30 prompt vectors are used, VPT-Deep outperforms FFT by 0.61% and LP by 33.61%.

Figure 9(a) presents the classification accuracy trends of VPT-Deep, FFT, and LP as the amount of training data in the downstream task increases, with 30 prompt vectors applied. As the downstream task training data increases, the performance of all three fine-tuning methods improves significantly. LP shows the poorest performance. With the increase in training data, the performance gap between VPT-Deep and FFT gradually decreases. When the data size in the downstream task is 480, the difference narrows to 0.61%. Despite this small margin, VPT-Deep offers major efficiency gains: for a ViT-Base encoder model with 85,803,270 parameters (327 MB), 30 prompts add only 276,480 parameters (1.07 MB), representing 0.322% of the total. Consequently, VPT-Deep needs to store just 1.07 MB, whereas FFT must save the entire 327 MB model. Figure 9(b) reports fine-tuning time. When the training data in the downstream task grows from 60 to 480 samples, FFT time rises from 0.84 h to 1.46 h, whereas VPT-Deep increases more modestly—from 0.51 h to 0.79 h. LP, tuning only 0.0716% of parameters, ranges from 0.45 h to 0.66 h, but achieves only 63.33 % accuracy. At 480 samples and 30 prompts, VPT-Deep reduces training time by 45% compared to FFT.

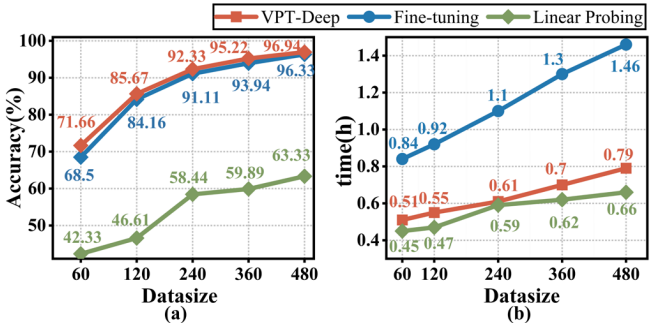

Fig. 9. (a) Classification accuracy and (b) fine-tuning time trends of VPT-Deep, FFT, and LP as the amount of training data in the downstream task increases, with 30 prompt vectors applied.

Figures 10(a)–(b) display the confusion matrices of VPT-Deep and FFT on downstream task testing sets. As defined in Table I, the six event classes correspond to gait data collected from Subjects I–III wearing slippers and sneakers. On the test set, VPT-Deep achieves a classification accuracy of 96.94%, outperforming FFT by 0.61%.

Based on the experimental results, we conducted a comprehensive comparison of VPT-Deep, FFT, and LP in terms of performance, resource consumption, and training time, as summarized in Table III. Evaluation metrics include the proportion of trainable parameters, parameter update scope, fine-tuning time, peak GPU memory usage, and classification accuracy on the downstream task. FFT updates all backbone parameters, resulting in the highest training parameter ratio (100%), the longest training time (1.46 hours), and the largest GPU memory usage (3511.76 MB). It achieves 96.33% accuracy on the downstream task. LP updates only the fully connected layer, with the lowest parameter ratio (0.0716%), the shortest training time (0.66 hours), and the smallest memory usage (776.40 MB), but yields a lower test accuracy of 63.33%. In contrast, VPT-Deep with 30 prompt vectors adjusts only 0.322% of parameters, requires 0.79 hours for training, and uses 2361.97 MB of memory, while achieving the highest test accuracy of 96.94%. Overall, VPT-Deep provides the best balance between performance and resource efficiency, making it the preferred choice for DAS-based event recognition systems.

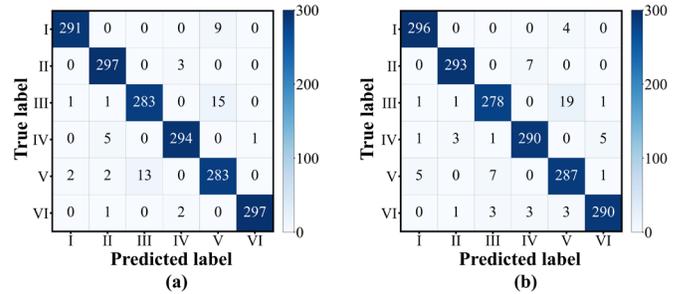

Fig.10. Confusion matrices of (a) VPT-Deep and (b) FFT on downstream task testing sets.

TABLE III
COMPARISON OF THE PERFORMANCE, COMPUTATIONAL COST, AND TRAINING TIME OF VPT-DEEP, FFT, AND LP IN DOWNSTREAM TASK

| Method | FFT | LP | VPT-Deep (p=30) |
|---|---|---|---|
| Tuned params | 100% | 0.0716% | 0.322% |
| Scope | Backbone | Linear | Input |
| Extra params | no | no | prompt vectors |
| Training time(h) | 1.46 | 0.66 | 0.79 |
| Peak GPU(MB) | 3511.76 | 776.40 | 2361.97 |
| Accuracy on downstream task | 96.33% | 63.33% | 96.94% |

*D. Water Pipe Leakage Detection Experiment*

To assess the generalizability and scalability of the MAEPD base model across diverse tasks and evaluate the efficiency of VPT, a DAS-based water pipe leakage detection platform was constructed in the laboratory, as depicted in Figure 11. The system comprises a water reservoir, a 750 W pump, and a U-shaped main pipeline. The pump draws water from the reservoir to simulate pipe leaks, with the U-shaped pipeline facilitating water circulation; water flows through the pipe and returns to the tank via a hose, forming a closed loop, while the pump is placed outdoors to minimize noise interference. The test pipe, a 32 mm radius PVC pipe, has a 5 mm leak hole at the detection point, fitted with a removable rubber stopper for quick switching between "no-leakage" and "leakage" conditions. A leakage branch channels leaked water back to the reservoir, ensuring safety and circulation efficiency. The single-mode fiber is spirally wound around the test pipe's outer wall, secured with epoxy resin and tape, enabling vibration detection for leakage identification. The DAS system collects 1-D raw signals at a 10 m spatial resolution and 2 kHz sampling rate, which are converted into 2D time-frequency images via short-time Fourier transform over a 2-second window. Four data categories—leakage, no-leakage, hammering (pipe struck with a hammer), and cutting (pipe cut with a saw)—were collected.

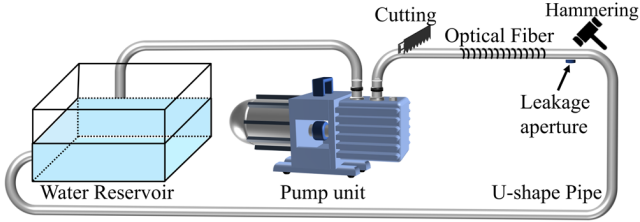

Fig.11. Schematic diagram of water pipe leakage detection experiment.

TABLE IV
EXPERIMENTAL CONFIGURATION FOR WATER PIPE LEAKAGE DETECTION

| Experimental configuration | VPT |
|---|---|
| Train datasize | 1600 |
| Test datasize | 1200 |
| The optimal VPT variant | VPT-Deep |
| The optimal prompt-insertion strategy | Top-Bottom: 1~12 |
| The optimal number of prompts | 30 |
| Tuned params | 0.322% (versus FFT) |
| Model memory footprint (MB) | 1.07 (FFT:327) |
| Training time(h) | 1.46(FFT: 3.62) |
| Peak GPU(MB) | 2375.67 (FFT:3197.39) |
| Top test accuracy on downstream task | 98.83%(FFT:98.50%) |

The system employs the same MAEPD foundation model as the backbone, integrated with VPT using the parameter settings from Table II. Table IV outlines the experimental setup for the water pipe leakage detection task, utilizing 1600 training samples (400 per category) and 1200 test samples (300 per category), all newly collected and excluded from the upstream dataset. Experiments were performed on an NVIDIA GeForce RTX 4080 Super platform. VPT-Deep emerged as the optimal variant for the leakage detection task, achieving peak performance with 30 prompt vectors inserted across all encoder layers. This configuration adjusts only 0.322% of parameters, requiring 1.07 MB of storage and 2375.67 MB of peak GPU memory (versus 3197.39 MB for FFT). Training time totals 1.46 hours (versus 3.63 hours for FFT), with a test accuracy of 98.83% (versus 98.50% for FFT). These findings confirm that the MAEPD model seamlessly adapts to water pipe leakage detection without structural modifications, demonstrating superior generalizability and scalability. Moreover, VPT's ability to surpass FFT by tuning only 0.322% of parameters highlights its efficiency.

## VII. CONCLUSION

This study developed and validated MAEPD, a pretrained foundational model tailored for DAS. Through self-supervised learning on a diverse dataset of DAS images from multiple application scenarios, MAEPD effectively captures universal semantic features of DAS signals. VPT was applied to fine-tune the pretrained MAEPD for downstream tasks. In gait recognition, the VPT-Deep approach achieved a classification accuracy of 96.94%, outperforming FFT at 96.33% and LP at 63.33%. While improving accuracy, VPT-Deep significantly reduced computational demands. Compared to FFT, which updates all parameters, VPT-Deep fine-tunes only 0.322% of parameters, cutting training time by 45% and substantially reducing GPU memory usage. The model's versatility was further confirmed through successful applications in scenarios such as pipeline leakage detection, highlighting its potential as a foundation for diverse downstream tasks. The proposed "MAEPD + VPT" framework offers an efficient, robust, and scalable solution to challenges in DAS, including limited AI model generalization and high training costs, effectively balancing performance and resource efficiency.

## ACKNOWLEDGMENT

This work was supported in part by National Natural Science Foundation of China (60907032, U23A2074); Natural Science Foundation of Zhejiang Province (LZ24F050008, LY20F050009); Open Fund of the State Key Laboratory of Advanced Optical Communication Systems and Networks (2020GZKF013); Horizontal projects of public institution (KY-H-20221007, KYY-HX-20210893).

## REFERENCES

[1] G. Tu, X. Zhang, Y. Zhang, F. Zhu, L. Xia and B. Nakarmi, "The development of an Φ-OTDR system for quantitative vibration measurement," *IEEE Photonics Technology Letters*, vol. 27, no. 12, pp. 1349-1352, 2015.
[2] J. Jiang, Z. Wang, Y. Wu, S. Lin, J. Xiong, Y. Chen and Y. Rao, "Coherent Kramers-Kronig receiver for Φ-OTDR," *IEEE Journal of Lightwave Technology*, vol. 37, no. 18, pp. 4799-4807, 2019.
[3] Z. He and Q. Liu, "Optical fiber distributed acoustic sensors: A Review," *IEEE Journal of Lightwave Technology*, vol. 39, no. 12, pp. 3671-3686, 2021.
[4] T. Yamate, G. Fujisawa, and T. Ikegami, "Optical sensors for the exploration of oil and gas," *IEEE Journal of Lightwave Technology,* vol. 35, no. 16, pp. 3538-3545, 2017.
[5] G. Naldrett, T. Parker, S. Shatalin, M. Mondanos, and M. Farhadiroushan, "High-resolution carina distributed acoustic fibre optic sensor for permanent reservoir monitoring and extending the reach into subsea fields," *First Break,* vol. 38, no. 2, pp. 71-76, 2020.
[6] I. Ashry, Y. Mao, B. Wang, F. Hveding, A. Bukhamsin and T. Khee, "A review of distributed fiber–optic sensing in the oil and gas industry," *IEEE Journal of Lightwave Technology,* vol. 40, no. 5, pp. 1407-1431, 2022.
[7] X. Li, Y. Zeng, N.D. Muchiri, W. Yan and Y. Feng, "The use of distributed acoustic sensing (DAS) in monitoring the integrity of cement-casing system," Journal of Petroleum Science and Engineering, vol. 208, pp. 109690-109702, 2022.
[8] M. Fontana, Á. F. García-Fernández, and S. Maskell, "Notch power detector for multiple vehicle trajectory estimation with distributed acoustic sensing," *Signal Processing,* vol. 232, pp. 109905, 2025.

[9] W. Huang, S. Chen, Y. Wu, R. Li, Y. Huang, X. Cao and Z. Li, "DAShip:A Large-Scale Annotated Dataset for Ship Detection Using Distributed Acoustic Sensing Technique," *IEEE Journal of Selected Topics in Applied Earth Observations and Remote Sensing,* vol. 18, pp. 4093-4107, 2025.

[10] Z. Li, J. Zhang, M. Wang, Y. Zhong, and F. Peng, "Fiber distributed acoustic sensing using convolutional long short-term memory network: a field test on high-speed railway intrusion detection," *Opt Express,* vol. 28, no. 3, pp. 2925-2938, 2020.

[11] R. Zhong, C. Y. Chiang, M. Jaber, "Intelligent vehicle monitoring: Distributed acoustic sensors enabled smart road infrastructure," *IEEE Internet of Things Journal*, vol. 12, no. 11, pp. 15211-15223, 2025.

[12] Y. Khacef, M. van den Ende, C. Richard, A. Ferrari and A. Sladen, "Precision Traffic Monitoring: Leveraging Distributed Acoustic Sensing and Deep Neural Networks," *IEEE Transactions on Intelligent Transportation Systems*, vol. 26, no. 6, pp. 7678-7689, 2025.

[13] S. Glubokovskikh, R. Pevzner, E. Sidenko, K. Tertyshnikov, B. Gurevich, S. Shatalin, A. Slunyaev and E. Pelinovsky, "Downhole distributed acoustic sensing provides insights into the structure of short‐period ocean-generated seismic wavefield," *Journal of Geophysical Research: Solid Earth,* vol. 126, no.12, 2021.

[14] J. Lindsey, T. CraigJonathan, B. Ajo-Franklin, "Illuminating seafloor faults and ocean dynamics with dark fiber distributed acoustic sensing," *Science*, vol. 366, pp. 1103-1107, 2025.

[15] A. Sladen, D. Rivet, J. Ampuero, L. Barros, Y. Hello, G. Calbris and P. Lamare, "Distributed sensing of earthquakes and ocean-solid earth interactions on seafloor telecom cables," *Nature Communications,* vol. 10, no. 1, pp.5777, 2019.

[16] A. Lellouch, R. Schultz, N.J. Lindsey1, B. L. Biondi, and W. L. Ellsworth, "Low‐magnitude seismicity with a downhole distributed acoustic sensing array—Examples from the FORGE geothermal experiment," *Journal of Geophysical Research: Solid Earth,* vol. 126 no.1, 2021.

[17] B. Han, H. Guan, J. Yao, Y. Rao, Z. Ran, and Y. Gong, "Distributed acoustic sensing with sensitivity-enhanced optical cable," *IEEE Sensors Journal,* vol. 21, no. 4, pp. 4644-4651, 2021.

[18] T. Nishimura, K. Emoto1, H. Nakahara, S. Miur, M. Yamamoto, S. Sugimura, A. Ishikawa and T. Kimura, "Source location of volcanic earthquakes and subsurface characterization using fiber-optic cable and distributed acoustic sensing system," *Sci Rep,* vol. 11, no. 1, pp.6319, 2021.

[19] V. Lai, M. Miller, C. Jiang, Y. Yang, F. Magrini, Z. Zhan and H. McQueen, "Passive seismic imaging of urban environments using distributed acoustic sensing: A case study from Melbourne, Australia," *The Seismic Record,* vol. 4, no. 4, pp. 308-317, 2024.

[20] C. Lyu, Z. Huo, X. Cheng, J. Jiang, A. Alimasi, and H. Liu, "Distributed optical fiber sensing intrusion pattern recognition based on GAF and CNN," *IEEE Journal of Lightwave Technology,* vol. 38, no. 15, pp. 4174-4182, 2020.

[21] M. Sun, M. Yu, P. Lv, A. Li, H. Wang and X. Zhang, "Man-made threat event recognition based on distributed optical fiber vibration sensing and SE-WaveNet," *IEEE Transactions on Instrumentation and Measurement,* vol. 70, pp. 1-11, 2021.

[22] S. Li, R. Peng, Z. Liu, and X. Liu, "Perimeter monitoring of urban buried pipeline threated by construction activities based on distributed fiber optic sensing and real-time object detection," *Opt. Express,* vol. 32, no. 2, pp. 2590-2606, 2024.

[23] Y. Wang, W. Zhou, B. Liu, J. Liu, Y. Hu, and Y. Fu, "GASF-ConvNeXt-TF algorithm for perimeter security disturbance identification based on distributed optical fiber sensing system," *IEEE Internet of Things Journal,* vol. 11, no. 10, pp. 17712-17726, 2024.

[24] Z. Zhong, T. Liu, H. Wu, J. Qiu, B. Du, G. Yin, and T. Zhu, "High-spatial-resolution distributed acoustic sensor based on the time-frequency-multiplexing OFDR," *Optics Letters,* vol.48, no. 21, pp. 5803-5806, 2023.

[25] D. Kandamali, X. Cao, M. Tian, Z. Jin, H. Dong and K. Yu, "Machine learning methods for identification and classification of events in φ-OTDR systems: a review," *Applied optics*, vol.61, no. 11, pp. 2975-2997, 2022.

[26] L. Shao, J. Zhang, X. Chen, D. Xu, H. Gu, Q. Mu, F. Yu, S. Liu, X. Shi, J. Sun, Z. Huang, X. Yang, H. Zhang, Y. Ma, H. Lu, C. Liu and C. Yu. "Artificial intelligence-driven distributed acoustic sensing technology and engineering application," *PhotoniX*, vol.6, no. 1. pp. 4, 2025.

[27] Y. Lei, X. Tang, M. Xia, T. Jiang, S. Liu, T. Li, D. Ba, and Y. Dong, "212-km ultra-long-distance hybrid Φ-OTDR/BOTDR based on remotely pumped optical amplification," *Optics Express,* vol. 33, no. 7. pp. 15827-15837, 2025.

[28] H. Wu, Y. Qian, W. Zhang, and C. Tang, "Feature extraction and identification in distributed optical-fiber vibration sensing system for oil pipeline safety monitoring," *Photonic Sensors,* vol. 7, no. 4, pp. 305-310, 2017.

[29] C. Lyu, Z. Huo, X. Cheng, J. Jiang, A. Alimasi, H. Liu, "Distributed optical fiber sensing intrusion pattern recognition based on GAF and CNN," *IEEE Journal of Lightwave Technology*, vol. 38, no. 15, pp. 4174-4182, 2020.

[30] Z. Sun, K. Liu, T. Xu, Y. Xu, W. Fang, K. Xue, "Intelligent sensing analysis using mel-time-frequency-imaging and deep learning for distributed fiber-optic vibration detection," *IEEE Sensors Journal*, vol. 22, no. 22, pp. 21933-21941, 2022.

[31] Z. Zhou, T. Xie, X. Wang, J. Qu, J. Shi, Y. Han, "Open-set recognition model for distributed fiber-optic vibration sensor based on hierarchical attention network with Openmax. *IEEE Sensors Journal* (*Early Access*), DOI: 10.1109/JSEN.2025.3570991, 2025.

[32] X. Liu H. Wu, Y. Wang, Y. Tu, Y. Sun, L. Liu, Y. Song and G. Yan, "A fast accurate attention-enhanced ResNet model for fiber-optic distributed acoustic sensor (DAS) signal recognition in complicated urban environments," *Photonics,* vol. 9, no. 10, pp.677, 2022.

[33] T. He, H. Li, S. Zhang, Z. Zeng, Z. Yan and Q. Sun, "A surveillance system for urban utility tunnel subject to third-party threats based on fiber-optic DAS and FPN-BiLSTM network," *IEEE Transactions on Instrumentation and Measurement,* vol. 73, pp. 1-9, 2024.

[34] H. Wu, X. Liu, X. Wang, Y. Wu, Y. Liu, Y. Wang and Y. Rao, "Multi-dimensional information extraction and utilization in smart fiber-optic distributed acoustic sensor (sDAS)," *IEEE Journal of Lightwave Technology*, vol. 42, no. 19, pp. 6967-6980, 2024.

[35] H. Lee, S. Lee, J. Kim, H. Jung, K. Yoon, S. Gandla, H. Park and S. Kim, "Stretchable array electromyography sensor with graph neural network for static and dynamic gestures recognition system," *npj Flex Electron*, vol. 7, no. 20, 2023.

[36] L. Duan, D. Xu and I. Tsang, "Domain Adaptation From Multiple Sources: A Domain-Dependent Regularization Approach," *IEEE Transactions on Neural Networks and Learning Systems*, vol. 23, no. 3, pp. 504-518, 2012.

[37] Y. Shi, Y. Li, Y. Zhang, Z. Zhuang, and T. Jiang, "An Easy Access Method for Event Recognition of Φ-OTDR Sensing System Based on Transfer Learning," *IEEE Journal of Lightwave Technology,* vol. 39, no. 13, pp. 4548-4555, 2021.

[38] N. He and J. Zhu, "A weighted partial domain adaptation for acoustic scene classification and its application in fiber optic security system," *IEEE Access*, vol. 9, pp. 2244-2250, 2021.

[39] S. Zhang, H. Li, C. Fan, Z. Zeng, C. Xiong, J. Wu, Z. Yan, D. Liu and Q. Sun, "Adaptive decentralized AI scheme for signal recognition of distributed sensor systems," *Opto-Electronic Advances,* vol. 7, no. 12, pp. 240119-240119, 2024.

[40] M. Bagherian, E. Sabeti, K. Wang, M. Sartor, Z. Nikolovska-Coleska and K. Najarian, "Machine learning approaches and databases for prediction of drug-target interaction: a survey paper," *Brief Bioinform*, vol.22, no.1, pp. 247-269, 2021.

[41] X. Hou, J. Liu, B. Xu, B. Liu, X. Chen, M. Llyas, I. Ellis, J. Garibaldi and G. Qiu, "Dual adaptive pyramid network for cross-stain histopathology image segmentation," *International Conference on Medical Image Computing and Computer-Assisted Intervention*, pp. 101-109, 2019.

[42] N. Fei, Z. Lu, Y. Gao, G. Yang, Y. Huo, J. Wen, H. Lu, R. Song, X. Gao, T. Xiang, H. Sun and J. Wen, "Towards artificial general intelligence via a multimodal foundation model," *Nat Commun*, vol.13, no. 3094, 2022.

[43] N. Ding, Y. Qin, G. Yang, F. Wei, Z. Yang, Y. Su, S. Hu, Y. Chen, C. Chan, W. Chen, J. Yi, W. Zhao, X. Wang, Z. Liu, H. Zheng, J. Chen, Y. Liu, J. Tang, J. Li and M. Sun, "Parameter-efficient fine-tuning of large-scale pre-trained language models," *Nat Mach Intell*, vol. 5, pp. 220–235, 2023.

[44] Y. Zhou, M. Chia, S. Wagner, M. Ayhan, D. Williamson, R. Struyven, T. Liu, M. Xu, M. Lozano, P. Woodward-Court, Y. Kihara, A. Altmann, A. Lee, E. Topl, A. Denniston, D. Alexander and P. Keane, "A foundation model for generalizable disease detection from retinal images," *Nature*, vol.622, pp. 156–163, 2023.

[45] R. Bommasani, D. Hudson, E. Adeli, et al, "On the opportunities and risks of foundation models," preprint arXiv:2108.07258, 2021.
[46] R. Chen, T. Ding, M. Lu, D. Williamson, G. Jaume, A. Zhang, D. Shao, M. Shaban, M. Williams, L. Oldenburg, L. Weishaupt, A. Vadya, L. Le, G. Gerber, S. Sahai, W. Williams and F. Mahmood, "Towards a general-purpose foundation model for computational pathology," *Nature medicine*, vol.30, no.3, pp. 850-862,2024.
[47] S. Khan, M. Naseer, M. Hayat, S. Zamir, F. Khan and M. Shah, "Transformers in vision: A survey," *ACM computing surveys*, vol. 54, no. 10, pp. 1-41, 2022.
[48] K. He, X. Chen, S. Xie, Y. Li, P. Dollár, and R. Girshick, "Masked autoencoders are scalable vision learners," *Proceedings of the IEEE/CVF conference on computer vision and pattern recognition*, pp. 16000-16009, 2022.
[49] X. Yu and C. Chen, "A robust operators' cognitive workload recognition method based on denoising masked autoencoder," *Knowledge-Based Systems*, vol. 301, pp. 112370, 2024.
[50] A. Faysal, M. Rostami, T. Boushine, R. Roshan, H. Wng and N. Muralidhar, "DenoMAE2. 0: Improving Denoising Masked Autoencoders by Classifying Local Patches," preprint arXiv:2502.18202, 2025.
[51] J. Duan, J. Chen and Z. He, "DAS-MAE: A self-supervised pre-training framework for universal and high-performance representation learning of distributed fiber-optic acoustic sensing," preprint arXiv:2506.04552, 2025.
[52] Z. Wang, L. Zhang, S. Wang, N. Xue, Fei. Peng, M. Fan, W. Sun, X. Qian, J. Rao and Y. Rao, "Coherent Phi-OTDR based on I/Q demodulation and homodyne detection," *Opt Express,* vol. 24, no. 2, pp. 853-858, 2016.
[53] Z. Spica, J. Ajo-Franklin, G. Beroza, B. Biondi, F. Cheng, B. Gaite, B. Luo, E. Martin, J. Shen, C. Thurber, L. Viens, H. Wang, A. Wuestefeld, H. Xiao and T. Zhu, "PubDAS: A PUBlic distributed acoustic sensing datasets repository for geosciences," *Seismological Society of America*, vol.94, no. 2, pp. 983-998, 2023.
[54] Q. Goestchel, W. Wilcock and S. Abadi, "Enhancing fin whale vocalizations in distributed acoustic sensing data," *The Journal of the Acoustical Society of America*, vol.157, no.5, pp. 3655-3666, 2025.
[55] T. Zhu, J. Shen and E. Martin, "Sensing earth and environment dynamics by telecommunication fiber-optic sensors: An urban experiment in Pennsylvania USA," *Solid Earth Discussions*, vol.12, no.1, pp. 219-235, 2021.
[56] X. Cao, Y. Su, Z. Jin and K. Yu, "An open dataset of φ-OTDR events with two classification models as baselines," *Results in Optics*, vol.10, pp. 100372, 2023.
[57] B. Lester, R. Ai-Rfou and N. Constant, "The Power of Scale for Parameter-Efficient Prompt Tuning," preprint arXiv:2104.08691, 2021.
[58] M. Jia, L. Tang, B. Chen, C. Cardie, S. Belongie, B. Hariharan and S. Lim, "Visual prompt tuning," *European Conference on Computer Vision*. pp.709-727, 2022.
[59] C. Han, Q. Wang, Y. Cui, W. Wang L. Huang, S. Qi and D. Liu. "Facing the elephant in the room: Visual prompt tuning or full finetuning?" preprint arXiv:2401.12902, 2024.
[60] C. Wan, L. Wang and V. Phoha, "A survey on gait recognition," *ACM Computing Surveys* (CSUR), vol. 51, no. 5, pp. 1-35, 2018.
[61] Y. Zhou, J. K. W. Yeoh, Y. E. Li, and W. Solihin, "Large scale indoor occupant tracking using distributed acoustic sensing and machine learning," *Building and Environment,* vol. 247, pp.111005, 2024.
[62] Y. Shi, Y. Zhang, S. Dai, L. Zhao, and C. Xu, "Footsteps detection and identification based on distributed optical fiber sensor and double-YOLO model," *Opt Express,* vol. 31, no. 25, pp. 41391-41405, 2023.
[63] N. Prottasha, U. Chowdhury, S. Mohanto, T. Nuzhat, A. Sami, M. Ali, M. Sobuj, H. Raman, M. Kowsher and O. Garibay, "PEFT A2Z: Parameter-efficient fine-tuning survey for large language and vision models," preprint arXiv:2504.14117, 2025.

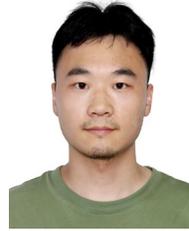

**Kun Gui** received the B.S. degree in Communications Engineering from the Henan Institute of Science and Technology in 2023. He is currently pursuing the M.S. degree in information and communication engineering with the School of Information Engineering, Zhejiang University of Technology, Hangzhou, China. His current research interests include distributed acoustic sensing and its related applications supported by artificial intelligence.

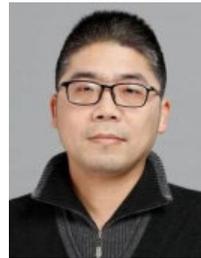

**Hongliang Ren** received the B.S. degree in optoelectronic technology and the M.S. degree in optical engineering from Zhengzhou University, Zhengzhou, China, in 2001 and 2004, respectively, and the Ph.D. degree in information and communication engineering from Shanghai Jiao Tong University, Shanghai, China, in 2008.

From 2008 to 2024, he served as a lecturer and an associate professor at the College of Information Engineering, Zhejiang University of Technology in Hangzhou, China. Since 2024, he has been serving as an associate professor at the Institute of Cyberspace Security, Zhejiang University of Technology. His research interests include optical sensing and computing, fiber optic communications, and associated applications of artificial intelligence.